# SHIELD: Securing Healthcare IoT with Efficient Machine Learning Techniques for Anomaly Detection

1st Mahek Desai
*Computer Science Department*
*California State University*
Northridge, USA
mahek.desai.849@my.csun.edu

2nd Apoorva Rumale
*Computer Science Department*
*California State University*
Northridge, USA
apoorva-sanjay.rumale.462@my.csun.edu

3rd Marjan Asadinia, IEEE Member
*Computer Science Department*
*California State University*
Northridge, USA
marjan.asadinia@csun.edu

*Abstract*—The integration of IoT devices in healthcare introduces significant security and reliability challenges, increasing susceptibility to cyber threats and operational anomalies. This study proposes a machine learning-driven framework for (1) detecting malicious cyberattacks and (2) identifying faulty device anomalies, leveraging a dataset of 200,000 records. Eight machine learning models are evaluated across three learning approaches: supervised learning (XGBoost, K-Nearest Neighbors (K- NN)), semi-supervised learning (Generative Adversarial Networks (GAN), Variational Autoencoders (VAE)), and unsupervised learning (One-Class Support Vector Machine (SVM), Isolation Forest, Graph Neural Networks (GNN), and Long Short-Term Memory (LSTM) Autoencoders). The comprehensive evaluation was conducted across multiple metrics like F1-score, precision, recall, accuracy, ROC-AUC, computational efficiency. XGBoost achieved 99% accuracy with minimal computational overhead (0.04s) for anomaly detection, while Isolation Forest balanced precision and recall effectively. LSTM Autoencoders underperformed with lower accuracy and higher latency. For attack detection, KNN achieved near-perfect precision, recall, and F1-score with the lowest computational cost (0.05s), followed by VAE at 97% accuracy. GAN showed the highest computational cost with lowest accuracy and ROC-AUC. These findings enhance IoT-enabled healthcare security through effective anomaly detection strategies. By improving early detection of cyber threats and device failures, this framework has the potential to prevent data breaches, minimize system downtime, and ensure the continuous and safe operation of medical devices, ultimately safeguarding patient health and trust in IoT-driven healthcare solutions.

## I. INTRODUCTION

The rapid integration of Internet of Things (IoT) devices in healthcare environments has revolutionized patient care through improved monitoring, diagnosis, and treatment efficiency. However, this increased connectivity introduces significant security and reliability challenges. Healthcare IoT devices, including continuous temperature monitors, blood pressure measurement systems, heart rate monitors, and battery-powered medical sensors, are vulnerable to both cyber threats and operational failures, potentially endangering patient safety, data privacy, and healthcare system integrity [1]. Cyberattacks targeting healthcare IoT can lead to unauthorized access to sensitive patient data, manipulation of device functionality, and disruption of critical healthcare services [2]. Similarly, faulty device anomalies can result in incorrect readings, treatment errors, and system downtime, compromising patient care [3]. According to recent studies, healthcare has become one of the most targeted sectors for cyberattacks, with a notable rise in data breaches in recent years [4], leading to significant financial losses per incident compared to other industries [5]. Additionally, device malfunctions in critical care settings pose serious risks, with an estimated 83% of medical devices vulnerable to exploitation [6], highlighting the need for robust anomaly detection systems.

Prior research has explored various machine learning (ML) approaches for IoT security, such as adaptive ML models (SVM, KNN, MLP, FusionNet) [7], clustering techniques (K-Means, K-Medoids) [8], and deep learning for medical anomaly detection [9], but many studies lacked real-time adaptability or comprehensive model comparisons. Other efforts integrated network intrusion detection with healthcare event monitoring using SVM and edge computing [10] or applied feature selection and Random Forest for cybersecurity anomaly detection [11], yet device-level anomalies remain underexplored. These gaps underscore the need for an integrated ML-driven framework addressing both cyber threats and faulty device anomalies in IoT-enabled healthcare environments.

To overcome existing limitations, we present SHIELD, a comprehensive framework for detecting both cyberattacks and faulty device anomalies in healthcare IoT environments. Using a dataset of 200,000 records from an ICU setup with patient monitoring sensors and control units, SHIELD employs a three-stage approach: (1) data preprocessing (cleaning, normalization, and feature engineering), (2) feature selection (ANOVA F-value, Mutual Information, Recursive Feature Elimination), and (3) model evaluation using supervised learning (XGBoost, KNN), semi-supervised learning (GAN, VAE), and unsupervised learning (One-Class SVM, Isolation Forest, GNN, LSTM Autoencoders). XGBoost demonstrated the highest performance for faulty device anomaly detection (99% accuracy, perfect precision and recall, 0.04 seconds). Isolation

Forest showed strong results with near perfect accuracy and recall. For cyberattack detection, KNN achieved near-perfect precision, recall, and F1-score, with a low computational cost (0.05 seconds), while GAN performed poorly (83% accuracy, 0.72 ROC-AUC). SHIELD's integrated approach ensures comprehensive protection against both operational and security threats across healthcare IoT ecosystems.

The remainder of this paper is structured as follows: Section II details our proposed methodology, including data collection, preprocessing techniques, feature selection methods, and model architectures. Section III provides a thorough evaluation of our framework, presenting comparative results across multiple performance metrics. Finally, Section IV concludes the paper with key insights and outlines future research directions in securing IoT healthcare systems.

## II. PROPOSED METHOD

This section provides a comprehensive overview of the proposed methodology. The workflow of our approach is illustrated in Figure 1.

### A. Data Collection

The dataset utilized in this study consists of 200,000 records and is divided into faulty device data and attack data, each capturing critical aspects of IoT-based Intensive Care Unit (ICU) conditions [12]. The data was obtained from an ICU setup featuring a two-bed capacity, where each bed is equipped with nine patient monitoring sensors and a Bedx-Control-Unit responsible for data aggregation and transmission.

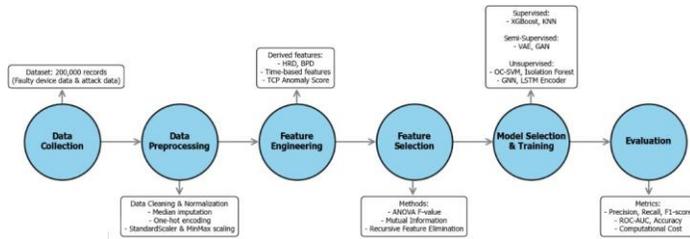

Fig. 1. SHIELD Overview

The faulty device data captures anomalies in medical device functionality, including temperature fluctuations, power failures, and erroneous sensor readings. This data consists of essential patient monitoring parameters. Additionally, control parameters such are included to ensure that deviations from expected values can be detected and analyzed. Table II summarizes all features.

The attack data documents malicious network activity targeting IoT-based medical infrastructure. It includes network traffic metadata, which provide insights into communication patterns within the network. Additionally, TCP and MQTT protocol flags are analyzed to detect network anomalies indicative of cyber threats such as MQTT protocol exploits and TCP flag anomalies. By capturing both operational failures and cybersecurity threats, this dataset provides a robust foundation for developing anomaly detection mechanisms tailored to IoT healthcare environments. Table I summarizes the IoT Medical Infrastructure Attack Data category and features.

TABLE I
STRUCTURED REPRESENTATION OF IoT MEDICAL INFRASTRUCTURE ATTACK DATA

| Category | Features |
|---|---|
| Network Traffic Metadata | frame.time_delta, frame.time_relative, frame.len |
| Source & Destination Information | ip.src, ip.dst, tcp.srcport, tcp.dstport, tcp.flags, mqtt.clientid, mqtt.msgtype |
| TCP Protocol Flags | tcp.flags.ack, tcp.flags.fin, tcp.flags.push, tcp.flags.reset, tcp.flags.syn |
| MQTT Protocol Flags | mqtt.conflag.qos, mqtt.retain, mqtt.topic |

TABLE II
STRUCTURED REPRESENTATION OF MEDICAL DEVICE DATA

| Category | Features |
|---|---|
| Patient Information | Patient ID, Timestamp |
| Sensor Data | Sensor ID, Sensor Type |
| Physiological Data | Temperature (°C), Systolic_BP (mmHg), Diastolic_BP (mmHg), Heart Rate (bpm), Battery Level (%) |
| Control Parameters | Target_Blood_Pressure, Target_Heart_Rate, Target Health Status |

### B. Data Preprocessing

Data preprocessing is crucial for enhancing the reliability and effectiveness of anomaly detection models. It involves data cleaning, normalization, and feature engineering to improve model performance and reduce dataset noise.

*1) Data Cleaning and Normalization:* To maintain the integrity of input features, missing values in numerical fields were handled through median imputation. This approach was chosen over mean imputation to mitigate the influence of extreme values or outliers, which are common in healthcare sensor readings. Furthermore, categorical variables such as Sensor_Type and ip.src were one-hot encoded, ensuring that each category was represented in a binary format suitable for machine learning models. This transformation prevents categorical variables from being misinterpreted as ordinal values, preserving the true nature of the data.

Different machine learning models require specific scaling techniques to standardize numerical features and improve learning efficiency. Traditional machine learning models, such as XGBoost, operate optimally with features scaled using StandardScaler. On the other hand, deep learning models, including Variational Autoencoders (VAE), Generative Adversarial Networks (GANs), and LSTM Autoencoders, benefit from MinMax scaling. This normalization method is particularly useful for gradient-based optimization, as it enhances convergence rates and stabilizes the training process. These preprocessing techniques collectively ensure that data is clean, well-structured, and appropriately scaled for subsequent model training.

*2) Feature Engineering:* To enhance anomaly detection, additional features were derived from sensor readings and network traffic logs, incorporating domain-specific knowledge. Heart Rate Deviation (HRD) and Blood Pressure Deviation (BPD) were calculated by measuring fluctuations in vitals over time, using the absolute difference between real-time measurements and the rolling mean within a predefined window. These deviations helped identify faulty sensor readings or health issues. Time-based features, like the hour of the day and day of the week, were extracted to identify temporal patterns in anomalies, such as increased cyberattacks during shift changes or sensor failures in prolonged operation.

Additionally, the TCP Anomaly Score was introduced by analyzing TCP flag distributions and network traffic behaviors to detect cyberattacks. Attributes like tcp.flags.ack, tcp.flags.push, tcp.flags.reset, and tcp.flags.syn were used to compute an anomaly score based on statistical deviations. A high score indicated a higher likelihood of malicious activity. These engineered features enriched the dataset, enabling the model to identify anomalies with greater precision, improving detection of both faulty medical devices and cyber threats in IoT healthcare environments.

These engineered features significantly enriched the dataset, providing deeper insights into potential anomalies beyond what raw data could offer. By incorporating domain knowledge into the preprocessing phase, the effectiveness of anomaly detection models was enhanced, ensuring robust performance in identifying faulty medical devices and cyber threats in IoT healthcare environments.

*C. Feature Selection*

Three feature selection methods were employed to identify the most significant features for anomaly detection in both cyberattack and faulty device detection tasks. The methods include ANOVA F-value, Mutual Information (MI), and Recursive Feature Elimination (RFE). The selected features were combined to form an optimal feature subset for each task, enhancing model performance and computational efficiency.

*1) ANOVA F-value Feature Selection:* ANOVA F-value feature selection was applied separately to attack detection and faulty device detection datasets to retain the most influential features. For cyberattack detection, key features such as tcp.connection.syn, frame.time_delta, ip.proto, frame.len, tcp.dstport, mqtt.ver, tcp.ack, tcp.flags.syn, and mqtt.msgtype were selected. Similarly, for faulty device detection, critical features such as Temperature (°C), Systolic_BP (mmHg), and Heart_Rate (bpm) were identified.

*2) Mutual Information (MI):* Unlike ANOVA, MI captures both linear and non-linear relationships, making it effective for complex datasets. This method was applied to both cyberattack detection and faulty device detection to rank features by their predictive power. For cyberattack detection, key features such as tcp.time_delta, mqtt.topic_len, tcp.flags.reset, tcp.flags.push, tcp.hdr_len, ip.ttl, and mqtt.dupflag were selected. For faulty device detection, critical features such as Diastolic_BP (mmHg), Heart_Rate (bpm), and Device_Battery_Level (%) were identified.

*3) Recursive Feature Elimination (RFE):* Recursive Feature Elimination (RFE) was applied separately to cyberattack and faulty device datasets to retain the most relevant features. For cyberattack detection, selected features included mqtt.qos, mqtt.retain, tcp.flags.ack, tcp.flags.fin, frame.time_relative, tcp. connection.rst, and mqtt.clientid_len. For faulty device detection, the selected features included Temperature (°C), Diastolic_BP (mmHg), and Heart_Rate (bpm).

*4) Integration of Feature Selection Techniques:* By integrating ANOVA F-value, Mutual Information, and RFE, the final dataset contained only the most relevant features for anomaly detection. For cyberattack detection, the final feature set consisted of 28 unique features, including network metadata, TCP flags, and MQTT protocol parameters. For faulty device detection, the final feature set retained key features consistently identified as important across different methods: Temperature (°C), Systolic_BP (mmHg), Diastolic_BP (mmHg), Heart_Rate (bpm), and Device_Battery_Level (%). This approach enhanced model accuracy, reduced computation time, and improved real-time performance.

*D. Model Selection & Training*

Machine learning plays a crucial role in both faulty device detection in healthcare IoT and cyberattack detection. These problems require different learning paradigms, ranging from supervised learning, where labeled data is available, to unsupervised learning, where the goal is to detect anomalies without predefined labels. Additionally, semi-supervised learning is useful in cases where only a small portion of labeled data is available. The implementation of eight distinct models spanning supervised (XGBoost, KNN), semi-supervised (GAN, VAE), and unsupervised (One-Class SVM, Isolation Forest, GNN, LSTM Autoencoder) learning paradigms represents a deliberate methodological choice rather than a conventional comparative study. This comprehensive approach mitigates overfitting risk by ensuring that anomaly detection remains consistent across fundamentally different algorithmic foundations, particularly crucial given our substantial 200,000 record dataset.

TABLE III
MODEL SPECIFICATIONS FOR FAULTY DEVICE DETECTION

| Model | Hyperparameters |
|---|---|
| XGBoost | Learning rate: 0.1, Max depth: 6, Evaluation metric: logloss |
| KNN | k=5, Distance metric: Euclidean |
| GAN | Latent space: 10, Epochs: 100, Batch size: 64 |
| VAE | Latent space: 2, Epochs: 100, Batch size: 32 |
| OC-SVM | nu=0.2, Kernel: RBF |
| Isolation Forest | Contamination: 0.2 |
| GNN | Hidden channels: 16, Epochs: 100, Learning rate: 0.01 |
| LSTM Autoencoder | LSTM units: 64, Epochs: 100, Batch size: 32 |

Each modeling paradigm addresses specific data characteristic challenges present in healthcare IoT: supervised models leverage known patterns, semi-supervised approaches handle limited labeled anomalies, and unsupervised methods detect novel threats. The consistent performance across all eight models provides robust evidence that detected anomalies represent genuine device faults or cybersecurity threats rather than statistical artifacts or model-specific biases—a critical consideration in healthcare environments where both false negatives and false positives carry significant consequences. The following sections describe the models used for each task, their architectures, and their implementations.

**(a) Supervised Learning**

*1) XGBoost:* For faulty device detection, XGBoost was implemented as a supervised classification approach. The implementation began with a stratified train-test split of scaled feature data, allocating 70% for training and 30% for testing while preserving the original anomaly distribution. The XGBoost classifier was configured with hyperparameters as specified in Table III. This gradient boosting framework leveraged sequential tree building to iteratively reduce classification errors. After training, the model was applied to the entire dataset to generate anomaly predictions. The computational cost was measured through precise timing of prediction phases.

In the healthcare cybersecurity application, XGBoost was implemented with similar configurations but employed a different methodological approach. Rather than using a train-test split, the model was trained on the entire labeled dataset, reflecting a scenario where some attack patterns had been previously identified and labeled. This supervised learning approach enabled the model to learn specific patterns associated with known cyberattack vectors in healthcare systems.

*2) K-Nearest Neighbors (KNN):* For faulty device detection, the K-Nearest Neighbors (KNN) algorithm was implemented as a supervised classification approach utilizing the same train-test split established for the XGBoost model. The KNN classifier was configured with hyperparameters as detailed in Table III. After training on the labeled subset, the model was applied to the entire dataset to generate anomaly predictions.

In the healthcare cybersecurity application, the KNN implementation maintained the same parameter configuration but employed a different methodological approach. The model was trained on the complete labeled dataset, allowing it to learn from the full spectrum of normal and attack patterns. Comprehensive evaluation metrics were calculated, including precision, recall, F1-score, accuracy, and ROC-AUC.

**(b) Semi-Supervised Learning**

*3) Generative Adversarial Networks (GAN):* For faulty device detection, a Generative Adversarial Network (GAN) architecture was implemented with complementary generator and discriminator networks. The model was configured with hyperparameters as listed in Table III. The training process involved alternating between discriminator and generator training phases.

For anomaly detection, noise vectors were used to generate synthetic device data, and the mean absolute error between real and generated samples served as the anomaly score. Observations exceeding the 80th percentile threshold of reconstruction errors were classified as anomalies.

In the healthcare cybersecurity implementation, the GAN was configured with a more sophisticated training protocol as detailed in Table IV. A critical methodological distinction in this implementation was the anomaly detection approach, which leveraged the discriminator's confidence scores rather than reconstruction error.

TABLE IV
MODEL SPECIFICATIONS FOR HEALTHCARE CYBERSECURITY

| Model | Hyperparameters |
|---|---|
| XGBoost | Learning rate: 0.1, Max depth: 6, Evaluation metric: logloss |
| KNN | k=5, Distance metric: Euclidean |
| GAN | Latent space: 10, Epochs: 1000, Batch size: 64 |
| VAE | Latent space: 10, Epochs: 200, Batch size: 32 |
| OC-SVM | nu=0.1, Kernel: RBF, Gamma: 0.1 |
| Isolation Forest | Contamination: 0.1 |
| GNN | Hidden channels: 32, Epochs: 100, Learning rate: 0.01 |
| LSTM Autoencoder | LSTM units: 64, Epochs: 200, Batch size: 32 |

*4) Variational Autoencoder (VAE):* In the faulty device detection implementation, a Variational Autoencoder (VAE) was architected with a dimensionality reduction approach, compressing input data of arbitrary dimension into a 2-dimensional latent space. The model was trained using hyperparameters as specified in Table III. Anomaly detection leveraged reconstruction error, with observations exceeding the 80th percentile threshold classified as anomalies.

The healthcare cybersecurity VAE implementation featured a more sophisticated architecture encapsulated within a custom model class. The dimensionality of the latent space was increased to 10, and the model was trained using hyperparameters as detailed in Table IV.

**(c) Unsupervised Learning**

*5) One-Class SVM:* For faulty device detection, a One-Class Support Vector Machine (OC-SVM) was implemented with hyperparameters as listed in Table III. This unsupervised approach was trained on augmented and scaled feature data to establish a decision boundary around normal operational patterns.

In the healthcare cybersecurity application, the One-Class SVM implementation was refined with domain-specific considerations. The model was exclusively trained on normal network traffic data using hyperparameters as detailed in Table IV.

*6) Isolation Forest:* For faulty device detection, the model was configured with hyperparameters as specified in Table III. The algorithm was applied to augmented and scaled feature data, with predictions mapped to a binary classification where -1 represented anomalies and 1 represented normal observations.

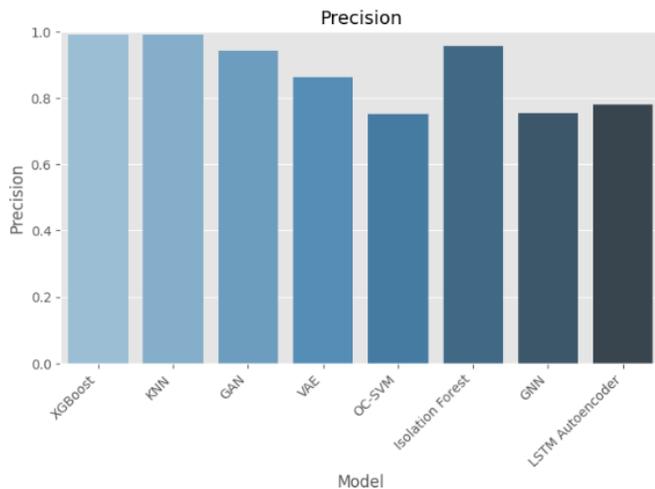

Fig. 2. Anomaly Detection Models Precision

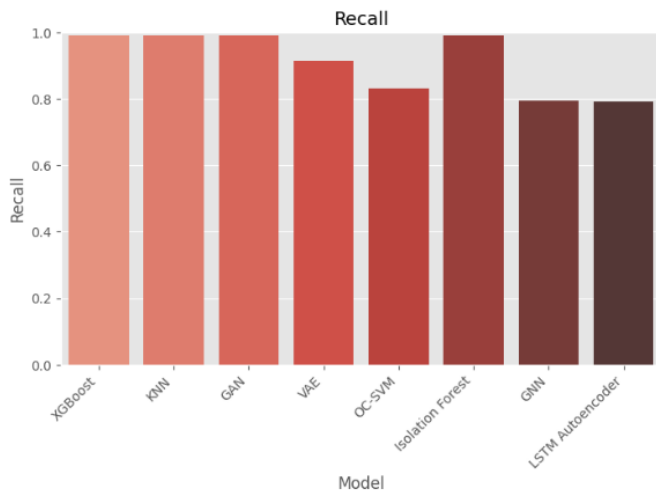

Fig. 3. Anomaly Detection Models Recall

In the healthcare IoT cybersecurity application, the Isolation Forest was configured with a reduced contamination parameter as detailed in Table IV.

*7) Graph Neural Networks (GNN):* For faulty device detection, a Graph Neural Network (GNN) architecture was implemented to leverage relational information between data points in the feature space. The model was configured with hyperparameters as listed in Table III. Anomaly detection utilized reconstruction error, with observations exceeding the 80th percentile threshold classified as anomalies.

In the healthcare cybersecurity implementation, the GNN approach was adapted to a supervised classification framework. The model architecture was modified to perform binary classification, with hyperparameters as detailed in Table IV.

*8) LSTM Autoencoder:* In the faulty device detection implementation, an LSTM Autoencoder model was configured with hyperparameters as specified in Table III. Anomaly detection was performed by calculating the reconstruction error (MSE) between original and reconstructed sequences.

For healthcare cybersecurity applications, a more sophisticated LSTM Autoencoder implementation was deployed with additional preprocessing steps. The model was trained using hyperparameters as detailed in Table IV.

## III. EVALUATION

The evaluation of anomaly detection models for Faulty Device Detection and Healthcare Cyberattack Detection relies on six key metrics: accuracy, precision, recall, F1-score, ROC-AUC, and computational cost (time to detect anomaly). Accuracy measures the overall correctness of the model, while precision focuses on the reliability of anomaly predictions, reducing false positives. Recall ensures the model detects as many anomalies as possible, minimizing false negatives, which is critical in healthcare systems.

The F1-score balances precision and recall, providing a single metric for imbalanced datasets. ROC-AUC evaluates the model's ability to distinguish between normal and anomalous

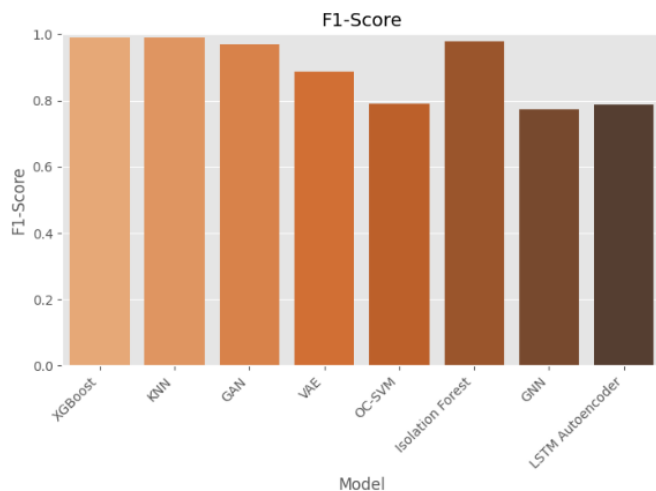

Fig. 4. Anomaly Detection Models F1-Score

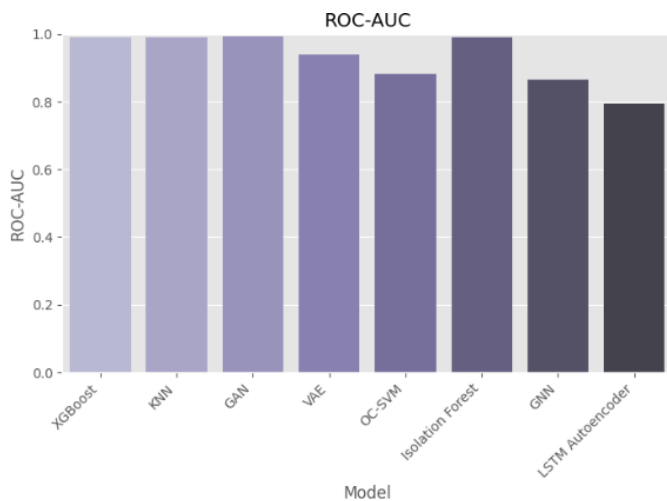

Fig. 5. Anomaly Detection Models ROC-AUC

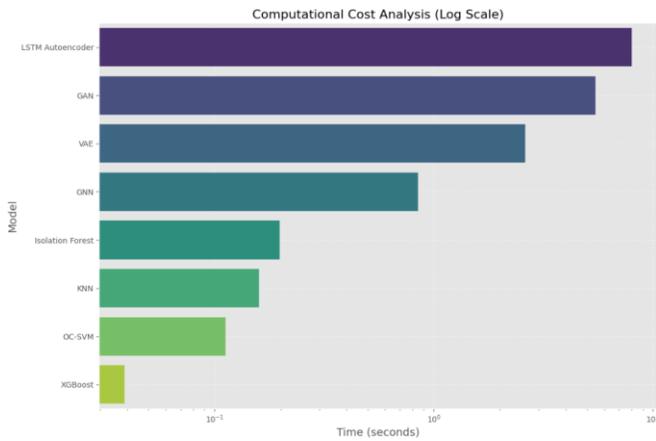

Fig. 6. Anomaly Detection Models Computational Cost (Log Scale)

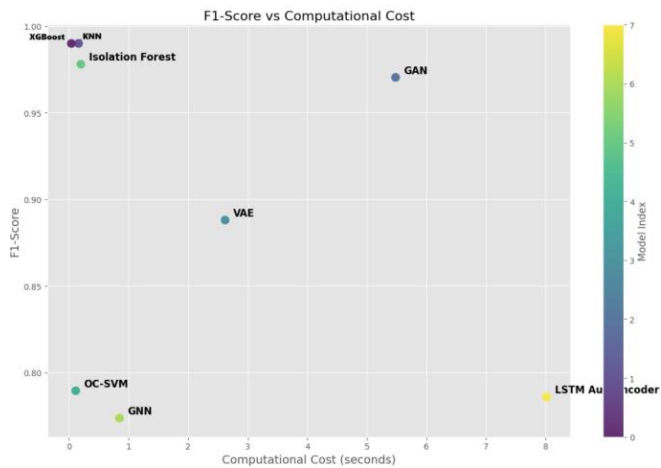

Fig. 7. Anomaly Detection Models F1-Score VS Computational Cost

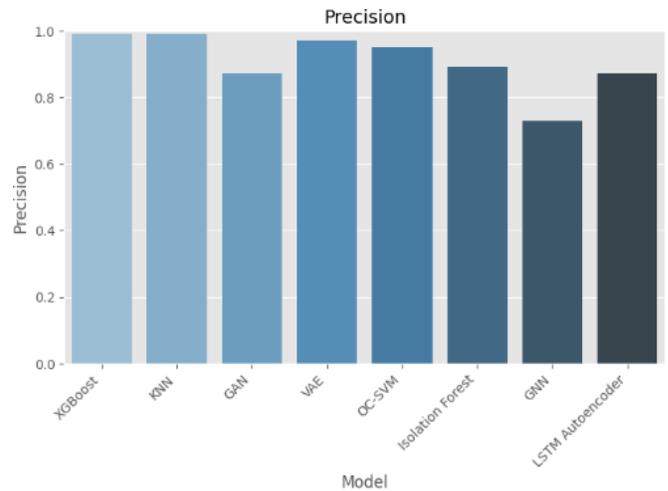

Fig. 8. Cyberattack Detection Models Precision

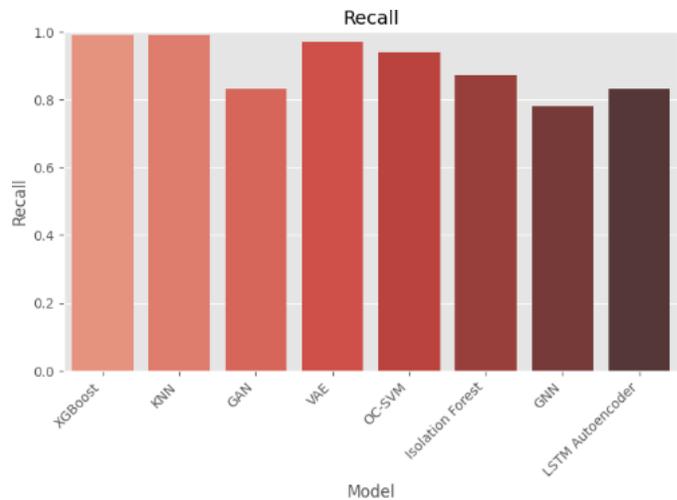

Fig. 9. Cyberattack Detection Models Recall

behaviors across thresholds, making it robust for imbalanced data.

Computational cost measures the time required for anomaly detection, ensuring real-time applicability. These metrics collectively ensure the models are accurate, reliable, and practical for deployment in real-world scenarios, where timely and precise anomaly detection is essential.

*1) Faulty Device Anomaly Detection Results:* In the Faulty Device Anomaly Detection Model, XGBoost and KNN emerged as the most effective and efficient models, outperforming all other approaches in terms of both predictive performance and computational efficiency. These models achieved top-tier results across multiple metrics, as shown in Figures 2 to 7, including accuracy, precision, recall, F1-score, and ROC-AUC, while also maintaining exceptionally low computational costs. Their ability to deliver high-quality results quickly makes them ideal choices for real-time anomaly detection in resource-constrained environments.

On the other hand, Isolation Forest and GAN also performed well, achieving near-perfect accuracy and recall, with impressive ROC-AUC scores (Figure 5). However, GAN was the least computationally efficient when compared to other models, with a significantly high computational cost (Figure 6), which could be a limiting factor in time-sensitive applications. Despite this, its strong overall performance makes it a viable option when training time is less critical.

VAE delivered robust results with a balanced performance across precision, recall, and ROC-AUC. Although it didn't match the top performers in overall effectiveness, it still maintained a respectable accuracy of. One-Class SVM, while offering moderate accuracy and an F1-score of 0.7894, proved to be a more practical choice for simpler systems that don't require the highest levels of precision. Notably, it had a very low computational cost of 0.1122, making it a lightweight option for quick anomaly detection.

In contrast, LSTM Autoencoder and GNN struggled to achieve high accuracy, precision, and ROC-AUC scores, with LSTM Autoencoder reaching only in ROC-AUC and an accuracy of. GNN, despite having a decent recall, had a lower

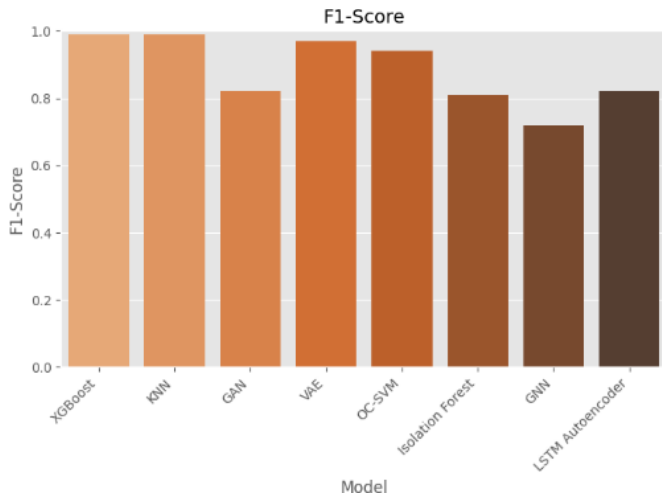

Fig. 10. Cyberattack Detection Models F1-Score

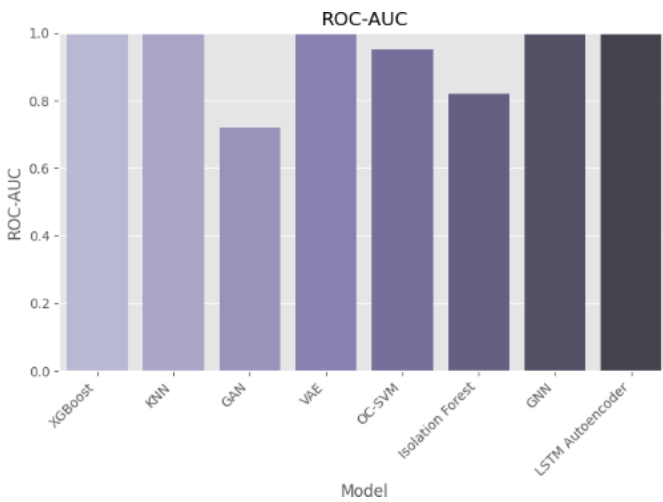

Fig. 11. Cyberattack Detection Models ROC-AUC

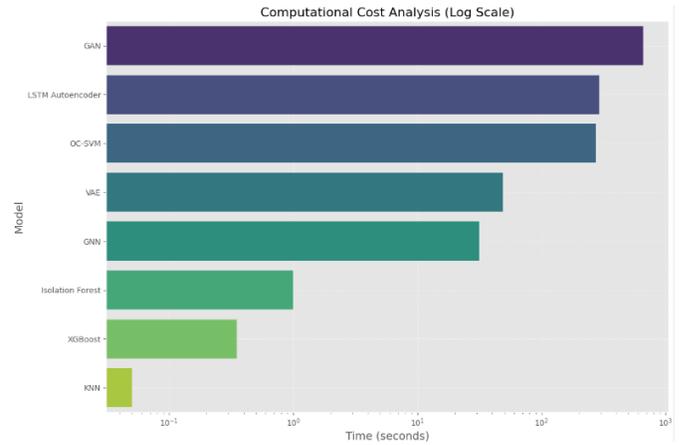

Fig. 12. Cyberattack Detection Models Computational Cost (Log Scale)

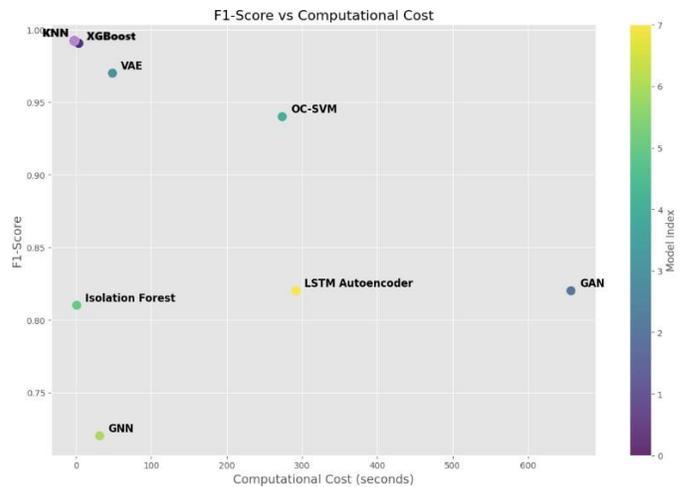

Fig. 13. Cyberattack Detection Models F1-Score VS Computational Cost

F1-score and accuracy. Moreover, their computational costs were significantly higher, with LSTM Autoencoder consuming 8.0163 and GNN, making them less efficient for the task (Figure 7). While these models are capable of modeling complex relationships in data, their higher computational costs and lower performance placed them behind the other methods.

Overall, XGBoost and KNN were the clear frontrunners, offering the best combination of accuracy and efficiency for detecting faulty devices. Even the lower-performing models still provided reasonably strong results, demonstrating that all tested approaches have potential applications depending on specific constraints such as computational cost and precision requirements.

*2) Healthcare Cyberattack Anomaly Detection Results:*
In the Healthcare Cyberattack Anomaly Detection Model, XGBoost and KNN emerged as the top-performing models, excelling across key metrics such as precision, recall, F1-score, accuracy, and ROC-AUC, while also maintaining minimal computational costs. This makes them ideal for real-time anomaly detection where both speed and accuracy are crucial. These models demonstrated a perfect balance of high performance and efficiency, enabling them to effectively identify cyberattacks while being resource-efficient as shown in Figures 8 to 13.

VAE and Isolation Forest also showed strong performance, with high precision (Figure 8), recall (Figure 9), and ROC-AUC scores (Figure 11). However, Isolation Forest had a slightly lower ROC-AUC, which could limit its effectiveness in detecting more nuanced anomalies. VAE offered a more balanced performance, providing reliable results across all metrics with an accuracy of 97%, but still fell short of the top performers.

One-Class SVM showed solid precision, recall, and F1-score (Figure 10), but suffered from high computational costs (Figure 12), making it less suitable for time-sensitive applications. LSTM Autoencoder, despite having a very high ROC-AUC, struggled with lower precision and recall and was the slowest model with a computational cost of 292.127s, significantly impacting its efficiency for real-time detection.

GAN and GNN presented mixed results, with GAN having the lowest accuracy and ROC-AUC, suggesting limited effectiveness for this task. Meanwhile, GNN, despite achieving a very high ROC-AUC, faced challenges with lower precision and recall, making it less effective for anomaly detection. Overall, XGBoost and KNN proved to be the most reliable and efficient models, providing the best combination of performance and speed for healthcare cyberattack detection. VAE and One-Class SVM offered competitive alternatives based on specific needs, while other models like LSTM Autoencoder, GAN, and GNN showed limitations in terms of both accuracy and computational efficiency (Figure 13).

## IV. CONCLUSION

The rapid adoption of IoT devices in healthcare has advanced patient care but also introduced significant security and reliability risks. Effective anomaly detection systems are essential to safeguard patient safety, data privacy, and healthcare system integrity. In this paper, we proposed SHIELD, a comprehensive framework designed to address these challenges. Through the use of a robust dataset and a multi-stage approach incorporating data preprocessing, feature selection, and diverse machine learning models, SHIELD effectively detects both cyberattacks and faulty device anomalies.

Our experimental results demonstrated that supervised learning models consistently outperformed other approaches, with XGBoost achieving superior performance for faulty device anomaly detection (99% accuracy, perfect precision and recall) and minimal computational overhead (0.04 seconds). Similarly, KNN excelled in cyberattack detection with near-perfect precision, recall, and F1-score (99%) at low computational cost (0.05 seconds). These results highlight supervised learning's effectiveness in handling labeled healthcare IoT data where anomaly patterns are well-defined. While semi-supervised (VAE: 97% accuracy) and unsupervised methods (Isolation Forest: 99% recall) showed competitive performance in specific scenarios, supervised models proved most reliable for real-time deployment. Notably, GAN underperformed among semi-supervised techniques (83% accuracy, ROC-AUC 0.72), suggesting limitations in adversarial training for healthcare IoT security. By combining these techniques, SHIELD offers a holistic solution to the security and reliability issues inherent in IoT-enabled healthcare environments, with supervised learning emerging as the optimal paradigm for mission-critical applications.

One of the greatest advantages of the SHIELD framework is its scalability, making it adaptable to a wide range of healthcare environments, from small clinics to large hospital networks. Thanks to its computationally efficient models, such as XGBoost and KNN, SHIELD can perform real-time anomaly detection even on low-power edge devices like wearable health monitors and IoT-connected medical sensors. Its modular design also allows for seamless integration with cloud-based security solutions, enabling large-scale anomaly detection across multiple healthcare facilities, regardless of location. SHIELD's scalability can be further enhanced by integrating federated learning for collaborative model training while preserving patient privacy. Additionally, time-series models and reinforcement learning would improve adaptability against evolving cyber threats and operational anomalies. Real-time streaming analytics and low-latency decision-making will optimize efficiency, making SHIELD ideal for high-demand healthcare networks. As smart healthcare systems advance, SHIELD will play a crucial role in securing and strengthening next-generation medical infrastructure.

## V. ACKNOWLEDGMENT

This work was supported by the National Science Foundation (NSF) under Grant No. [2318553]. We gratefully acknowledge the NSF for providing the financial support and resources necessary to conduct this research. The views expressed in this paper are those of the authors and do not necessarily reflect those of the NSF.